\documentclass[conference]{IEEEtran}
\IEEEoverridecommandlockouts
\usepackage{cite}
\usepackage{amsmath,amssymb,amsfonts}
\usepackage{algorithmic}
\usepackage{algorithm}
\usepackage{graphicx}
\usepackage{textcomp}
\usepackage{xcolor}
\def\BibTeX{{\rm B\kern-.05em{\sc i\kern-.025em b}\kern-.08em
    T\kern-.1667em\lower.7ex\hbox{E}\kern-.125emX}}
\begin{document}

\title{Efficient Multi-Crop Saliency Partitioning for Automatic Image Cropping\\}

\author{
	\IEEEauthorblockN{Andrew Hamara, Andrew C. Freeman}
\IEEEauthorblockA{Baylor University, 
Waco, TX, United States \\
\{andrew\_hamara1, andrew\_freeman\}@baylor.edu}
}


\maketitle

\begin{abstract}
Automatic image cropping aims to extract the most visually salient regions while preserving essential composition elements. Traditional saliency-aware cropping methods optimize a single bounding box, making them ineffective for applications requiring multiple disjoint crops. In this work, we extend the Fixed Aspect Ratio Cropping algorithm to efficiently extract multiple non-overlapping crops in linear time. Our approach dynamically adjusts attention thresholds and removes selected crops from consideration without recomputing the entire saliency map. We discuss qualitative results and introduce the potential for future datasets and benchmarks.
\end{abstract}


\section{Introduction}
Image cropping is the task of reducing the size of an image while preserving its most visually significant regions. As manual image cropping is a rather menial task, automatic image cropping (AIC) has seen significant exploration in the computer vision and multimedia communities. Many solutions to AIC have been proposed, ranging from traditional handcrafted feature-based methods \cite{handcrafted_1, handcrafted_2, handcrafted_3, handcrafted_4, handcrafted_5, handcrafted_6} to more modern, data-driven approaches \cite{learned_1, learned_2, learned_3, learned_4}. In either case, existing methods focus on single-crop extraction, optimizing either for a fixed aspect ratio bounding box based on saliency maps \cite{saliency_1, saliency_2, saliency_3, saliency_4, saliency_5, saliency_6}, or for aesthetic-based scoring \cite{aesthetics_1, aesthetics_2, aesthetics_3}. 

As pointed out by Ardizonne et al. \cite{saliency_4}, a single crop may be insufficient when multiple disjoint salient regions exist in an image, such as in panoramas, high-resolution images, and group photos. In such cases, optimizing for a single region may lead to loss of important visual information or only marginally reduce the size of the input image.

To address these challenges, we propose an efficient multi-crop saliency partitioning algorithm that extends the Fixed Aspect Ratio Cropping method proposed by Chen et al. \cite{chen2016automatic} to extract $k$ disjoint crops in linear time. The remainder of this paper is structured as follows: Section 2 reviews related work on automatic cropping and multi-region selection. Section 3 describes our proposed method and its computational benefits. Section 4 presents qualitative results, and Section 5 concludes the paper with directions for future research.

\section{Related Work}

There are several ways to logically distinguish existing AIC approaches. First, there is the distinction between handcrafted feature-based methods and data-driven approaches. Handcrafted methods rely on predefined rules such as the rule of thirds, edge detection, or symmetry, while data-driven methods learn cropping preferences from large-scale datasets. Second, AIC methods can be categorized based on whether they prioritize saliency-based cropping, which focuses on identifying visually important regions, or aesthetic-based scoring, which aims to improve image composition.

\subsection{Traditional Handcrafted Approaches}

The earliest attempts at AIC relied on well-known photography heuristics to suggest regions of interest. While computationally efficient, these approaches lacked generalization when compared to their learned counterparts. Additionally, they struggled with multi-object images and scenes with complex foreground-background relationships, as they did not take semantic content into account.

\subsection{Data-Driven Deep Learning Approaches}

Recent advances in deep learning-based cropping have demonstrated superior performance by training models on large-scale human-annotated datasets \cite{ni2022composition, learned_2, learned_5, learned_1, learned_3, learned_4}. Most methods use convolutional neural networks (CNNs) to predict the best crop based on image aesthetics \cite{learned_1, learned_2, learned_3, learned_4}. Reinforcement learning (RL) has also been applied with comparable results \cite{learned_5}.

Despite the noted advantages, deep learning-based cropping methods often require large computational resources, making them impractical for real-time or mobile applications. Furthermore, these models are fundamentally designed for single-crop selection, as they optimize a single bounding box per image. Since these models are not trained for multi-region cropping, an additional partitioning strategy is required to segment the image into disjoint subregions before they can be applied effectively.

\subsection{Saliency-Based Cropping}

Saliency-based image cropping stems from early vision studies on human gaze, which sought to understand how viewers naturally focus on specific regions of an image \cite{gaze_1, gaze_2}. With advancements in computer vision, saliency detection shifted from handcrafted rules and gaze estimation to data-driven models, leveraging large-scale eye-tracking datasets and deep learning techniques to predict where human observers are likely to focus. The integration of saliency prediction into automatic cropping was a natural extension, as it provided a numeric way to determine which areas of an image should be preserved. Many cropping algorithms optimize a single bounding box based on these saliency maps, assuming that the most attended region is also the most compositionally sound \cite{chen2016automatic, ni2022composition}.

However, despite its general effectiveness, researchers working on aesthetics-based AIC have noted that preserving a specific threshold of saliency does not necessarily imply that the optimal compositional crop was obtained \cite{saliency_alone_insufficient_1, saliency_alone_insufficient_2}.

\subsection{Aesthetic-Based Cropping}

Aesthetic-based cropping methods prioritize composition quality over raw saliency with the aim to produce visually pleasing crops that align with photography principles. These techniques ultimately attempt to mimic the decision-making process of professional photographers when selecting an optimal crop.

A number of large-scale image aesthetics assessment (IAA) datasets exist, where human annotators either score images on a scale (e.g, from 1-10) \cite{ava, aadb}, classify images as having ``good'' or ``bad'' composition \cite{cuhkpq}, or, given two images, selecting the more aesthetically pleasing of the two. Given the ability of a model to predict the aesthetic quality of an image, current aesthetic approaches to AIC simply generate candidate crops and compare them in a pairwise fashion to predict the optimal subregion \cite{learned_2}.

\subsection{Automatic Cropping of Multiple Regions}

The only prior work that directly attempts multi-region cropping, as proposed by Chen et al. \cite{chen2016automatic}, iteratively segments the image into disjoint subregions and selects the smallest valid crop in each subregion that satisfies a saliency threshold. The worst-case complexity is $O(m^2n^2)$ for $k = 2$ crops, and for $k > 2$, it scales even higher due to the explosion of possible partitions. Our work builds upon this approach but dramatically improves its efficiency by reducing runtime complexity to linear growth in the number of crops, $k$.

However, as noted previously, there is a dispute in the literature as to the definition of an optimal crop. The most closely related work \cite{chen2016automatic} defines the optimal crop as the smallest possible subregion that meets a fixed saliency threshold. Rather than treating these bounding boxes as fixed ground truth, we introduce an adjusted algorithm that first partitions the input image into disjoint subregions. These subregions can then be processed individually using existing single-crop models to extract $k$ visually meaningful crops.

\section{Proposed Approach}

\subsection{Problem Formulation}

Given an input image $I$ of dimensions $m \times n$, we define its corresponding attention map $G$, where $G(i, j)$ represents the attention value (ranging from 0-1) at pixel $(i, j)$. Larger values in $G$ indicate higher visual importance.

The objective of our algorithm is to find $k$ non-overlapping saliency partitions that properly distribute the attention values of $G$, such that the resulting partitions can be processed with a downstream image cropping model. To efficiently compute the summed saliency within a given partition, we utilize integral attention maps.

Let $G^+$ denote the integral attention map, allowing for rapid summation of any rectangular subregion in constant time. Formally, the integral map is defined as:

\[
G^+(i, j) = \sum_{k=1}^{i} \sum_{l=1}^{j} G(k, l),
\]

where $G(k, l)$ is the attention value at pixel $(k, l)$. Additionally, we define a column integral map with the cumulative attention sum of each column: 

\[
G^+_c(i, j) = \sum_{k=1}^{i} G(k, j).
\]

Both maps can be computed in $O(mn)$ time as a preprocessing step, significantly reducing the complexity of summing the attention within candidate regions.

\subsection{Multi-Region Partitioning Strategy}

The goal of our partitioning strategy is to extract disjoint saliency regions while minimizing computation. Our method follows an iterative selection process based on a dynamically updated saliency threshold, $\tau_k$. 

At each step, the algorithm computes the total remaining attention in $G$, denoted as $S_r$, and selects the smallest valid region satisfying $\tau_k$. In practice, this threshold is set slightly above $S_r / (k - n)$, where $n$ is the number of crops already selected. This assumes that saliency is unevenly distributed across the map, enabling the algorithm to prioritize regions with the highest relative attention. Once a valid region is identified, its attention values are suppressed to zero to prevent re-selection in subsequent iterations.

After extracting $k$ bounding boxes, the next step is to determine the partition boundaries that divide the image into disjoint regions. We compute partition lines by finding the midpoints between adjacent bounding boxes along their nearest edges. If two bounding boxes are horizontally adjacent, the partition line is drawn at the vertical midpoint between them; if they are vertically adjacent, the partition is drawn at the horizontal midpoint. This ensures that the selected regions are distinctly separated, allowing for further independent processing of each partition.

With the partitioning strategy established, we now present the algorithmic implementation, detailing the iterative region selection process in pseudocode.

\subsection{Partitioning Algorithm Implementation}

We leverage the $maxSubarrayFL$ algorithm, a linear-time method for extracting the best subarray of a fixed width that meets a given saliency threshold. A complete explanation of this algorithm, along with its theoretical guarantees, can be found in Chen et al. \cite{chen2016automatic}. For our purposes, it suffices to understand that $maxSubarrayFL$ identifies the optimal subregion with worst-case $O(n)$ complexity.

\begin{algorithm}
\caption{Multi-Region Saliency Partitioning}
\label{alg:multi_crop}
\textbf{Input:} $G$: A non-negative saliency map of size $m \times n$, $r$: Aspect ratio, $k$: Number of partitions \\
\textbf{Output:} $R_k$: A set of $k$ disjoint bounding boxes, $B_k$: Partition boundary lines
\begin{algorithmic}[1]
\STATE Compute integral maps $G^+$ and $G^+_c$
\STATE Initialize empty lists $R_k$ and $B_k$
\STATE $S \gets G^+(m, n)$, $S_r \gets S$ \COMMENT{Total and remaining saliency}
\FOR{$i \gets k$ \textbf{to} $1$}
    \STATE $\tau_i \gets \tau_k$
    \STATE $T \gets \tau_i \cdot S_r$
    \STATE $i_1 \gets 1$, $i_2 \gets 1$, $S_{\min} \gets -1$
    \STATE $i^*, j^*, w^*, h^* \gets 0, 0, \infty, \infty$ \COMMENT{Best crop variables}
    \WHILE{$i_2 \leq m$}
        \STATE $h_0 \gets i_2 - i_1 + 1$, $w_0 \gets \lceil h_0 \times r \rceil$
        \IF{$w_0 > n$}
            \STATE $i_1 \gets i_1 + 1$
        \ELSE
            \STATE $\hat{a} \gets G^+_c(i_2, :) - G^+_c(i_1 - 1, :)$
            \STATE $j_1, S_0 \gets \text{maxSubarrayFL}(\hat{a}, w_0, T)$
            \IF{$j_1 > 0$}
                \IF{$w_0 h_0 < w^* h^* \vee (w_0 h_0 = w^* h^* \wedge S_0 > S_{\min})$}
                    \STATE $i^*, j^*, w^*, h^* \gets i_1, j_1, w_0, h_0$
                    \STATE $S_{\min} \gets S_0$
                \ENDIF
                \STATE $i_1 \gets i_1 + 1$
            \ELSE
                \STATE $i_2 \gets i_2 + 1$
            \ENDIF
        \ENDIF
    \ENDWHILE
    \IF{$w^* \neq \infty$}
        \STATE Append $(i^*, j^*, i^* + h^* - 1, j^* + w^* - 1)$ to $R_k$
        \STATE Suppress $G$: $G[i^*:i^* + h^* - 1, j^*:j^* + w^* - 1] \gets 0$
        \STATE Update $S_r \gets \text{sum}(G)$, increase $\tau$
    \ENDIF
\ENDFOR
\STATE Sort $R_k$ by coordinates (left-to-right, top-to-bottom)
\FOR{$i \gets 1$ \textbf{to} $|R_k| - 1$}
    \STATE Compute boundary between $R_k[i]$ and $R_k[i+1]$
    \STATE Append boundary to $B_k$
\ENDFOR
\STATE \textbf{return} $R_k, B_k$
\end{algorithmic}
\end{algorithm}

\section{Qualitative Results}

\begin{figure}[t]
    \centering
    \begin{tabular}{cc}
        \includegraphics[width=0.45\linewidth]{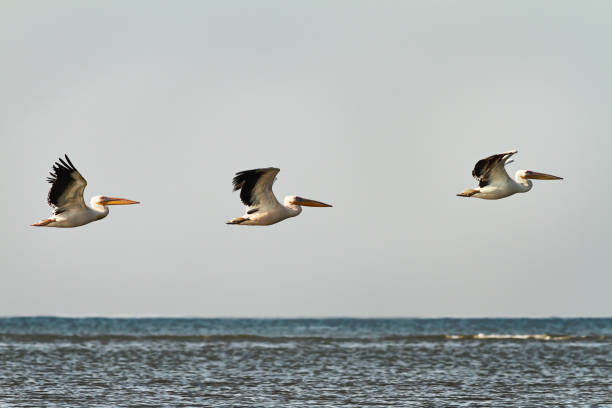} &  
        \includegraphics[width=0.45\linewidth]{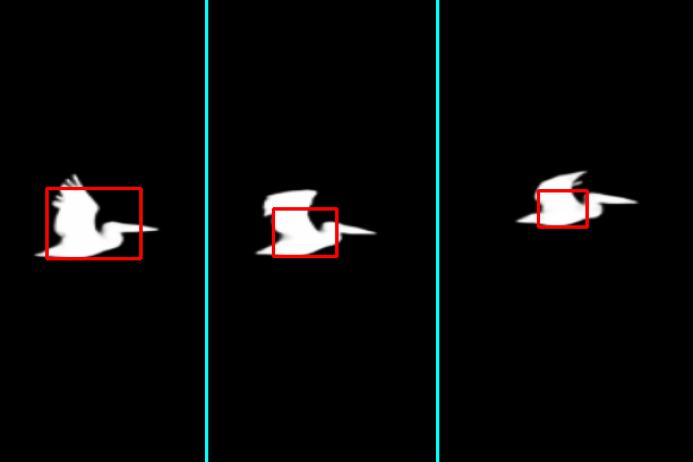} \\
        (a) Original Image 1 & (b) Partitioned Image 1 \\
        \includegraphics[width=0.45\linewidth]{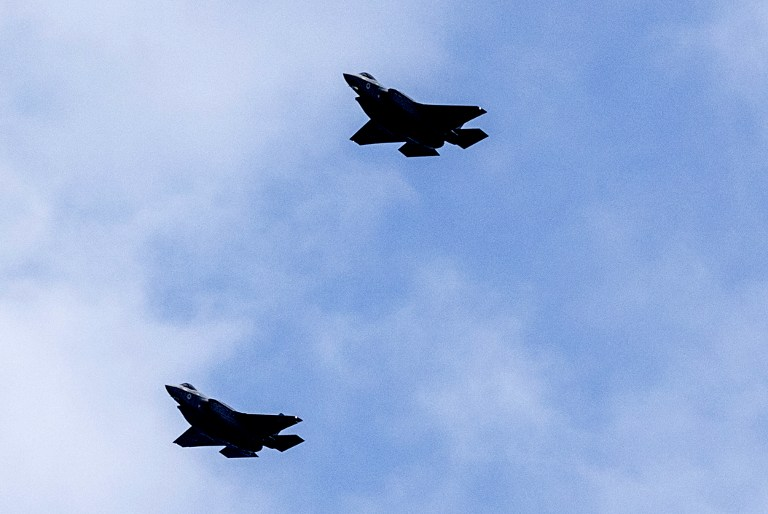} &  
        \includegraphics[width=0.45\linewidth]{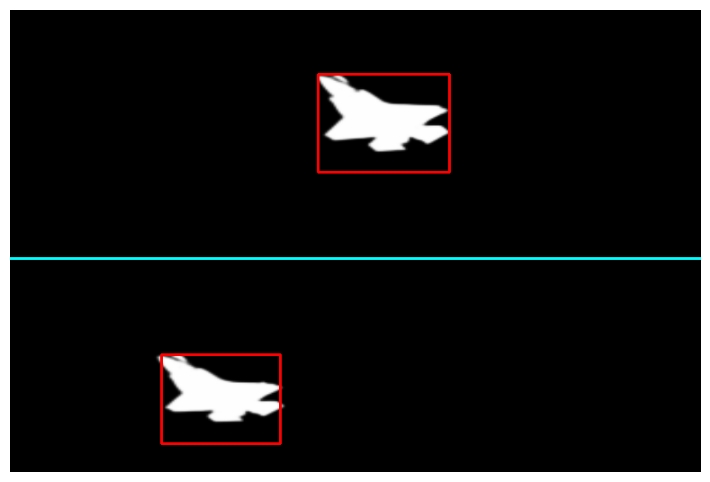} \\
        (c) Original Image 2 & (d) Partitioned Image 2 \\
    \end{tabular}
    \caption{Comparison of original images and their corresponding multi-crop partitions, overlaid on their corresponding saliency maps. Cyan lines indicate partition boundaries, and red bounding boxes represent cropped regions.}
    \label{fig:multi_crop_results}
\end{figure}

To assess the effectiveness of our multi-region saliency partitioning method, we present qualitative results demonstrating how the algorithm extracts multiple disjoint high-saliency regions from images with both horizontal and vertical structure.

Figure \ref{fig:multi_crop_results} presents example images before and after applying our partitioning algorithm. The bounding boxes, while included for clarity, serve only as intermediate outputs for partitioning. These partitions define the disjoint image segments, allowing for individual processing by pre-trained AIC models without redundant region overlap or excessive cropping beyond salient content.

By enforcing spatial separation, our approach enables a modular cropping pipeline: each partitioned segment can be fed independently into a downstream cropping model, ensuring that aesthetically and semantically relevant compositions are maintained. This structured pre-processing step is particularly beneficial in multi-object scene understanding, adaptive image retargeting, and thumbnail selection, where traditional single-crop approaches often fail to generalize.

\section{Conclusion and Future Work}

While our qualitative results demonstrate the potential of multi-region saliency partitioning for automatic image cropping, a quantitative evaluation is necessary to validate it. Such an evaluation would require a benchmark dataset with disjoint bounding boxes annotated for each input image. However, to the best of our knowledge, no publicly available dataset provides multi-crop ground truth annotations. Constructing such a dataset would enable comprehensive benchmarking and allow future research to thoroughly explore multi-region aesthetic cropping.

Our approach also introduces certain limitations. First, it assumes a reasonably accurate saliency map. If saliency is poorly aligned with compositional features, our algorithm will likely fail to correctly partition the image. Additionally, the performance gains achieved by our approach come at the cost of increased sensitivity to the saliency threshold parameter $\tau$.

Future work will focus on curating a labeled multi-crop dataset and developing quantitative metrics for evaluating saliency partitioning strategies. With such a dataset, we can systematically compare our method against competitive models, assess its impact on downstream AIC pipelines, and refine the algorithm based on objective performance feedback.

\bibliographystyle{IEEEtran}
\bibliography{references}

\end{document}